%% file: lenu.tex
\pdfoutput=1
\documentclass[sigconf]{acmart} 
\makeatletter
\renewcommand\@formatdoi[1]{\ignorespaces}
\makeatother
\renewcommand\footnotetextcopyrightpermission[1]{}
\usepackage{CJKutf8}
\usepackage{colortbl}

\acmDOI{XXXXXXX.XXXXXXX}

\setcopyright{none}
\begin{document}

\title{Transformer-based Entity Legal Form Classification}

\author{Alexander Arimond}
  \affiliation{%
    \institution{Sociovestix Labs}
    \country{Edinburgh, Scotland, UK}
  }
  \email{alexander.arimond@sociovestix.com}

\author{Mauro Molteni}
  \affiliation{%
    \institution{Sociovestix Labs}
    \country{Edinburgh, Scotland, UK}
  }
  \email{mauro.molteni@sociovestix.com}

\author{Dominik Jany}
  \affiliation{%
    \institution{GLEIF}
    \country{Basel, Switzerland}
  }
  \email{dominik.jany@gleif.org}

\author{Zornitsa Manolova}
  \affiliation{%
    \institution{GLEIF}
    \country{Basel, Switzerland}
  }
  \email{zornitsa.manolova@gleif.org}

\author{Damian Borth}
  \affiliation{%
    \institution{University of St. Gallen (HSG)}
    \country{St. Gallen, Switzerland}
  }
  \email{damian.borth@unisg.ch}

\author{Andreas G.F. Hoepner}
  \affiliation{%
    \institution{University College Dublin}
    \country{Dublin, Ireland}
  }
  \email{andreas.hoepner@ucd.ie}

\renewcommand{\shortauthors}{Arimond, et al.}

\begin{abstract}
    \input{00_abstract.tex}

\end{abstract}


\maketitle

\input{01_introduction.tex}

\input{02_related_work.tex}

\input{03_methodology.tex}

\input{04_experiments.tex}

\input{05_conclusion.tex}

\bibliographystyle{ACM-Reference-Format}
\bibliography{references}

\end{document}

%% file: 00_abstract.tex
We propose the application of Transformer-based language models for classifying entity legal forms from raw legal entity names.
Specifically, we employ various BERT variants and compare their performance against multiple traditional baselines. 
Our evaluation encompasses a substantial subset of freely available Legal Entity Identifier (LEI) data, comprising over 1.1 million legal entities from 30 different legal jurisdictions. 
The ground truth labels for classification per jurisdiction are taken from the Entity Legal Form (ELF) code standard (ISO 20275).

Our findings demonstrate that pre-trained BERT variants outperform traditional text classification approaches in terms of F1 score, while also performing comparably well in the Macro F1 Score. 
Moreover, the validity of our proposal is supported by the outcome of third-party expert reviews conducted in ten selected jurisdictions.

This study highlights the significant potential of Transformer-based models in advancing data standardization and data integration.
The presented approaches can greatly benefit financial institutions, corporations, governments and other organizations in assessing business relationships, 
understanding risk exposure, and promoting effective governance.

%% file: 01_introduction.tex
\section{Introduction}

\input{011_table_inconsistent_legal_form_representations.tex}
Identifying and understanding an entity's legal form is a crucial component in many financial and business related processes.
The legal form and structure of corporations can inform how to conduct transactions effectively and serves as a risk indicator. 
Legal forms indicate the owner's responsibility in case of financial problems. 
Depending on the legal form, an entity's owner might only be liable to a limited amount of money. 
This may determine whether an entity tends to make risky decisions which in turn will affect investors and creditors \cite{jensen1976, campos2003}.
Automating the process of identifying an entity's legal form can therefore lower risk, create transparency and increase operational efficiency 
by enabling straight-through processing (STP) capabilities.

Knowing the entity legal form also has high value in data integration and deduplication tasks \cite{eurostat2010}. 
Depending on the task at hand, entity data needs to be linked across several sources, while often times there is no common unique identifier available. 
As a consequence, entity linkage methods typically aim to identify similarities between legal names, addresses and other features of the entities.
Due to differing representations of such features across sources, this methodology usually results in many ambiguous entity mapping pairs and thus requires additional manual effort. 
Introducing the entity's legal form via standardized data items adds an additional layer of confidence into such entity linkage tasks, enabling 
robust mapping pairs across multiple datasets \cite{kruse2021developing}, as each entity can only have one single legal form.

The wide range of entity legal forms existing within and between different jurisdictions has made it challenging for organizations 
to categorize and structure this information effectively. 
This task becomes even more difficult due to the similarities in types and textual representation of these legal forms across jurisdictions. 
For instance, as one can see in the examples (a) and (b) in Table \ref{table:inconsistentlegalform}, different jurisdictions can have their distinct versions of a "Limited Liability Company" (LLC), 
where each one operates under its specific legal framework.
It is essential to note that even if multiple jurisdictions, like US-Delaware and US-New York, have LLCs, 
they are distinct entity forms. The complexity therefore lies in handling the diversity of entity legal forms across jurisdictions and the need to distinguish between seemingly 
similar legal structures. 

Furthermore, the representation of legal forms within entity names is influenced by the cultural and linguistic context of the 
jurisdiction and the specific legislation used to manage registrant information. 
Expert interviews have highlighted that, for example, many entities in France are not legally obligated to include their legal form in their names. 

Additionally, entity legal forms of the same type may be inconsistently represented, especially across different data sources.
These discrepancies often manifest in different punctuation styles, as shown in the examples (d) and (e) in Table \ref{table:inconsistentlegalform}, 
or in the use of various forms of abbreviations as demonstrated by the entities (f), (g) and (h):
the legal form "Aktiebolag" can be written out in full at the end of the name or represented by the abbreviation "AB".
Moreover, it can also appear at the beginning of the entity's name as "Aktiebolaget".
In Germany, according to its official business register of the German Federal Statistical Office 2018, of the names of the more than 9,500 entities 
that are stock corporations, 13\% include the term "Aktiengesellschaft" while 87\% include the abbreviation "AG". 
Similar figures result for the almost 700,000 private limited companies (Gesellschaft mit beschränkter Haftung, GmbH) \cite{LaurentESRB}.

Inconsistencies are further introduced by variations in the capitalization of individual characters or the entire name. 
Many business registers default to representing all information in uppercase letters, which adds to the complexity, as shown with examples (b) and (c) in Table \ref{table:inconsistentlegalform}.
Also, the intermingling of legal form and name elements as in example (i), the absence of any mention of the legal form like in example (j), 
or the usage of non-latin characters in examples (k) and (l),
exacerbates the challenge of accurately detecting legal forms, in particular if data users lack local domain expertise. 
In summary, the significant variability in representation presents considerable obstacles for any automated identification approach.

In this paper, we explore the application of machine learning and deep learning techniques to accurately classify entity legal forms. 
Our focus lies specifically on utilizing the Entity Legal Form (ELF) code (ISO 20275)\footnote{\label{ISOELF}https://www.iso.org/standard/67462.html}
as a fundamental basis for classification. 
The ELF code standard serves as a comprehensive solution for standardized legal form representation and is incorporated in the open LEI data, which 
encompasses over 2.4 million entities worldwide as of July 2023. 
Consequently, the LEI data and the ELF code standard provide an optimal data source for training classifiers dedicated to legal form detection.

We address the challenge of legal form classification by employing a novel approach using Transformer language models based on the BERT architecture in combination with standardized ELF codes. 
The results of the Transformer models are compared to a traditional Bag-of-Words setting.
To evaluate the performance of both Transformer and traditional approaches, a substantial subset of LEI data comprising over 1 million legal entities from 30 different legal jurisdictions is employed as the evaluation dataset.
Lastly, we conducted an expert review using 7,256 entities, further corroborating the plausibility of our findings.

\subsection{LEI Data}

As a basis for our work we use the open and freely available Global LEI Repository. 
As of July 2023, there are over 2.4 million entities listed in the LEI system. 
This global directory is centered on the Legal Entity Identifier (LEI). 
The LEI is a 20-digit alpha-numeric code based on the ISO 17442 standard\footnote{https://www.iso.org/standard/78829.html}. 
It connects to the key reference information that enables clear and unique identification of legal entities, and greatly enhances transparency by making available dependable and trusted data free of charge.
The reference data is maintained by so-called LEI issuers that are distributed across the globe. 
These LEI issuing organizations apply their local knowledge to verify the information of each LEI record. 
ELF code, legal name and jurisdiction are part of each LEI record\footnote{https://www.gleif.org/en/lei-data/access-and-use-lei-data}.
The data is easily accessible as csv, json, xml\footnote{https://www.gleif.org/en/lei-data/gleif-golden-copy/download-the-golden-copy/} and 
via a freely accessible API\footnote{https://www.gleif.org/en/lei-data/gleif-api}. 
The Global Legal Entity Identifier Foundation (GLEIF) ensures high data quality and offers transparent data quality reporting\footnote{https://www.gleif.org/en/lei-data/gleif-data-quality-management/quality-reports} \footnote{https://dq-dashboard.gleif.org/dashboard} 
and can therefore be considered a suitable source for our work.

\subsection{Entity Legal Forms (ELF) Code List}

The ELF code, established by the International Organization for Standardization (ISO)\footnotemark[1], 
is a unique 4-digit alpha-numeric code and serves as a comprehensive solution for standardized legal form representation. 
As of July 2023 there are 3,250 legal forms in 175 jurisdictions worldwide available\footnote{https://www.gleif.org/en/about-lei/code-lists/iso-20275-entity-legal-forms-code-list} 
in version 1.4.1 of the openly accessible ELF code list. 
GLEIF has been acting as the maintenance agency secretariat of the ELF code list since 2017, 
regularly introducing new legal forms and jurisdictions. 
By way of example, Table \ref{table:comparison} shows the ELF codes for US-Delaware for entities that have been revalidated recently\footnote{Entity.EntityStatus == "ACTIVE" and Registration.RegistrationStatus == "ISSUED"}. 
Each jurisdiction has its unique list of legal forms. Legal forms that appear to be similar across jurisdictions must be treated individually due to differing legislature within each jurisdiction. ELF code \texttt{8888} is a so-called reserved code that is used in case no specific legal form can be assigned. 
Notably, the legal forms are not equally distributed within US-Delaware. 
Similarly imbalanced legal form data can be observed in all jurisdictions.

\begin{table}[h]
\caption[]{Legal forms and their ELF codes in US-Delaware for entities in scope\footnotemark[9]}
\begin{tabular}{lll}

\toprule
Legal form name & ELF code & \#Entities  \\
\midrule

Commercial Bank & \texttt{9ASJ} & 2 \\
Corporation & \texttt{XTIQ} & 5,379 \\
Limited Liability Company & \texttt{HZEH} & 30,553 \\
Limited Liability Limited Partnership & \texttt{TGMR} & 44 \\
Limited Liability Partnership & \texttt{1HXP} & 64 \\
Limited Partnership & \texttt{T91T} & 9,707 \\
Non-deposit Trust Company & \texttt{MIPY} & 2 \\
Partnership & \texttt{QF4W} & 16 \\
Savings Bank & \texttt{JU79} & 0 \\
Statutory Trust & \texttt{4FSX} & 1,266 \\
Unincorporated Nonprofit Association& \texttt{12N6} & 1 \\
Legal form not yet in code list & \texttt{8888} & 7,118 \\

\bottomrule

\end{tabular}
\end{table}

Using this standardized and up-to-date list of entity legal forms, data users are enabled to uniquely identify the legal forms of entities around the globe without having any local knowledge about language or legislature. 
This benefits data users as they do not need to carry out costly legal name analysis. 
For instance, there are more than 500 legal form checks implemented in Deutsche Bundesbank's Financial Statement Data Pool on approximately 125,000 entities each year. 
These checks would not be necessary, if all entities were assigned with an ELF code \cite{LaurentESRB}. 
\\
\\
The remainder of this work is structured as follows: 
In Section 2, we provide an overview of related work. 
Section 3 follows with an explanation of our baseline setup with traditional Bag-of-Words, as well as a description of the methodology of using Transformers in the context of legal form classification.
The experimental setup and results are outlined in Section 4, while Section 5 showcases the results of an expert review.
In Section 6, the paper concludes with a summary of the work.

%% file: 011_table_inconsistent_legal_form_representations.tex
\begin{table*}[h!]
\caption{Examples of inconsistent legal form representations} \label{table:inconsistentlegalform}
\begin{tabular}{cllll}

\toprule
    &Legal name & Jurisdiction & Legal form name & ELF code \\
\midrule
    (a) & Dean Quarry Apartments LLC & US-NY & Limited Liability Company & \texttt{SDX0} \\
    (b) & RUBICON TECHNOLOGY MANAGEMENT L.L.C. & US-DE & Limited Liability Company & \texttt{HZEH} \\
    (c) & LOCKWOOD RIVERFRONT HOTEL, LLC & US-DE & Limited Liability Company & \texttt{HZEH} \\
    (d) & GIANT Weilerswist g21 GmbH & DE & Gesellschaft mit beschränkter Haftung & \texttt{2HBR} \\
    (e) & Selbstfahrer Union G.m.b.H. & DE & Gesellschaft mit beschränkter Haftung & \texttt{2HBR} \\
    (f) & Interproximal AB & SE & Aktiebolag & \texttt{XJHM} \\
    (g) & Konstlist i Heby Aktiebolag & SE & Aktiebolag & \texttt{XJHM} \\
    (h) & Aktiebolaget Clas Grönwalls Lantbrukstjänst ilkividation & SE & Aktiebolag & \texttt{XJHM} \\
    (i) & Infrastrukturentwicklungsgesellschaft Hilden mbH & DE & Gesellschaft mit beschränkter Haftung & \texttt{2HBR} \\
    (j) & Katholische Kirchengemeinde Maria Königin Lingen & DE & Körperschaft des öffentlichen Rechts & \texttt{SQKS} \\ 
    (k) & \begin{CJK*}{UTF8}{min}むつ小川原風力合同会社\end{CJK*} & JP & \begin{CJK*}{UTF8}{min}合同会社\end{CJK*} & \texttt{7QQ0} \\
    (l) & \begin{CJK*}{UTF8}{min}合同会社まつお\end{CJK*}  & JP & \begin{CJK*}{UTF8}{min}合同会社\end{CJK*} & \texttt{7QQ0} \\
\bottomrule

\label{table:inconsistentlegalform}
\end{tabular}
\end{table*}

%% file: 02_related_work.tex
\section{Related work}

In this section, we present our literature study and existing approaches to entity legal form classification, while 
highlighting similarities and differences to our work.

Schild et al. \cite{schild10f} presented a technical report that demonstrated record linkage on company entities using machine learning. 
They applied and evaluated their approach by matching seven distinct German datasets. 
Their matching approach involved comprehensive preprocessing, including legal form pattern extraction and standardization, 
achieved through regular expression matching. 
Despite impressive results (90\% coverage), creating meaningful regular expressions required local expertise in German legal forms and is difficult to scale
to non-german jurisdictions. 
In contrast, our proposed approach is more automated, with the underlying model autonomously identifying relevant tokens representing legal forms.

A similar rule-based approach is taken by the open-source Python library \textit{cleanco}\footnote{https://github.com/psolin/cleanco}. 
This library aims to detect and eliminate the legal form from a company's name by incorporating an extensive list of legal form abbreviations and commonly encountered suffixes.
According to the main author\footnote{https://paulsolin.com/2014/03/16/cleanco-a-python-module-for-checking-business-names/}, 
the rules were mainly distilled from Wikipedia articles\footnote{https://en.wikipedia.org/wiki/List\_of\_legal\_entity\_types\_by\_country}. 
The library's coverage encompasses 16 distinct entity types mapped to 66 countries. 
Notably, the library does not apply the concept of jurisdictions within countries. 
For instance, it does not differentiate individual states within the United States of America. 
Assigning the legal forms to their respective states is necessary as each state represents a unique jurisdiction in which similar sounding legal forms can have different legal implications. 
Also, the legal forms that are returned to the user do not represent the actual local name of the legal form but are instead translated into an English 
legal form that is assumed similar to the actual legal form. 
This can cause ambiguity as legal forms are unique within their respective jurisdictions and should therefore be represented in their local 
naming convention to avoid any misjudgement. 
In our work we avoid any ambiguity of legal forms by utilizing ELF codes that provide a unique link to the local name of the legal form.

Loster et al. \cite{loster2018dissecting} have worked on dissecting company names into various elements including the core name, 
legal form, personal names and location names. 
To address this task, they formalized the problem as a supervised sequence labelling task and evaluated 
a linear chain conditional random field classifier. 
Furthermore, they explored an extensive set of potential features on a manually annotated dataset consisting of 
1,500 randomly selected company names, primarily of German origin. 
For 1,484 out of the 1,500 annotated legal names they were able to annotate one part of the legal name as a representation 
of the entity's legal form. 
For recognition of the legal form tag, their approach achieved a 98\% F1-Score. 
We showcase in our work how Transformer based ELF code classification models can be utilized to attribute legal form relevance on token level. 

Kruse et al. \cite{kruse2021developing} explicitly tackle the problem of legal form classification in the context of data integration in the absence of identifier keys. 
In particular, they claim that extracting the legal form into a separate attribute (besides name, address and others) can enhance the performance of record linkage.
Their hybrid approach to legal form extraction - which they evaluated on a German-only subset - involves a rule-based component for identifying the legal form. 
We not only claim that this can be achieved through machine learning as presented in this work, 
but that its beneficial to learn statistical relationships in cases where names lack explicit representation of the corresponding legal form.

%% file: 03_methodology.tex
\section{Methodology}

In this section, we present how to identify entity legal forms by applying text classification techniques. 
We distinguish between a traditional approach, and a modern approach based on Transformer neural networks.

\subsection{Traditional Approach}

Text classification has been widely studied in the field of Natural Language Processing (NLP)
and has important applications like sentiment analysis, news categorization, email classification and spam filtering \cite{sebastiani2002machine, aggarwal2012survey}. 
In this regard, a wide variety of traditional machine learning techniques have been successfully applied for decades, 
such as Naive Bayes classifiers, Decision Tree classifiers, Random Forests and Support Vector Machines.

Our baseline approach to legal form classification follows a traditional text classification pipeline setup, 
consisting of pre-processing of input text, a feature selection step and the training of a classifier.

The \textbf{pre--processing} "cleans up" the input text to achieve a minimum degree of harmonization, 
which in turn eases subsequent processing and enhances classification performance.
For this, we compared two pre-processing approaches: 
\begin{itemize}
    \item[(a)] we apply a simple approach by transforming each input name string to lower case letters. This ensures a minimal degree of harmonization of the input names. This is the default setup.
    \item[(b)] we adopt a more extensive set of harmonization rules by following ideas for record linkage as presented in \cite{magerman2006data} and \cite{magnani2007study}.
This includes (1) converting the string to lower case, (2) replacing diacritics, (3) replacing multi-spaces, (4) removing double quotation marks, (5) replacing trailing non-alphanumeric characters, (6) correcting commas and periods, (7) applying purge rules and (8) replacing multi-spaces.
        Our purge step removes special characters like "-", "(", ")", ";", "/", "," with simple white spaces and converts " \& " and " + " to " and ". We denote this setup with \textbf{" + prep"}.
\end{itemize}

In the \textbf{feature selection} step, we chose to transform each input legal entity name into their \textit{Bag-of-Words} representation,
due to its simplicity for classification purposes \cite{aggarwal2012survey}. 
For this, we split the (pre-processed) legal name strings at their white spaces into their set of words. 
In typical document classification scenarios, each word gets assigned its frequency in the document. 
In contrast, since legal names are comparably short and therefore rare to have duplicate words,  
we decided to only count each word once by default, resulting in a binary representation vector:

\begin{equation} \label{eq:feat2}
X_{j} =
\begin{cases}
    1 & \text{if legal name contains word}  w_j \\
    0 & \text{otherwise} 
\end{cases}
\end{equation}

The words in the training corpus constitute the so called vocabulary. Its size is comparably large, as legal names by nature consist of very distinctive words.
This also means that the vector dimensionality is very high. Also, it should be noted that this kind of representation disregards information about the 
position of words in the input name. 

For \textbf{classification} we experimented with several methods, the first to mention is the \textit{Complement Naive Bayes (CNB)} algorithm, 
which is an adaptation of the standard \textit{Multinomial Naive Bayes} algorithm, which is particularly suited for imbalanced data sets \cite{rennie2003tackling}.
Furthermore, we applied a \textit{Decision Tree classifier} (DT), which divides the underlying data space with the use of different text features \cite{aggarwal2012survey}. 
For a given text instance, the most likely partition can then be assigned by applying a set of learned decision rules, which is then used for the purposes of classification.
We also applied \textit{Random Forest} (RF) \cite{breiman2001random}, which - as an ensemble method - fits a number of decision tree classifiers on sub-samples of the data and uses averaging to improve predictive
accuracy and avoid over-fitting. As a last method we have used \textit{Support Vector Machines} (SVM), which attempt to partition the data with the use of 
non-linear delineations between the different classes \cite{aggarwal2012survey}.
For all presented traditional classifiers, we make use of the implementations of the python machine learning library \textit{scikit-learn}\footnote{https://scikit-learn.org} in its version 1.3.0.
We do not apply any modifications to the default parameter settings as defined by the library.

\subsection{Transformers}

In recent years, with the rise of deep learning, and in particular the Transformer model \cite{vaswani2017attention}, 
a new class of neural network language models have surpassed traditional machine learning--based approaches in many text classification benchmarks \cite{minaee2021deep}.

As our second approach to entity legal form classification, we chose to explore the usage of Transformer language models. 
One of the most popular of these models are Bidirectional Encoder Representations from Transformers (BERT) \cite{devlin2018bert} by Google,
which can be pre-trained on large corpuses of unlabelled text. 
A pre-trained BERT model can be easily fine-tuned to various NLP tasks - including text classification - and as such enables low-resource tasks to benefit from deep architectures. 
Google\footnote{https://github.com/google-research/bert} itself, as well as HuggingFace \cite{wolf2019huggingface} provide many variants of BERT, 
including the original “base” English and multilingual versions, to be further used and fine-tuned by the community.
In this study, we apply BERT to the entity legal form classification problem, evaluating different variants of pre-trained BERT models.

Language models come along with pre-trained tokenizers specifically tied to the model. BERT uses WordPiece, which tokenizes the input string into sub-word units, resulting
in a vocabulary size of 30,000 tokens. By using sub-word units, it can effectively deal with rare words (as, for instance, given in legal names), 
and allows for a good balance between the flexibility of single characters and the efficiency of full words \cite{wu2016google}.
We omitted any custom pre-processing or tokenization in favor of following the end-to-end processing as defined by the BERT variants\footnote{Some models may incorporate specific pre--processing steps, such as uncased versions of BERT that strip our accent markers.}.

In contrast to the Bag-of-Words model, which loses information about the position of words, BERT is a sequence model.
The output of the model is an embedding vector that is supposed to capture the whole sequence of tokens, including positional context information. 
When fine-tuning the model for classification, the sequence embedding serves as input to a classification head (usually a single layer of randomly initialized weights). 
During training, not only the classification head, but also the pre-trained weights within the model are tuned to the task.
We will outline in section 4 how these characteristics can be helpful to identify those tokens of a legal name that are relevant for the legal form attribution. 

Due to its availability and strong performance, a zoo of pre-trained and fine-tuned variations of BERT has emerged, 
in particular with the goal to account for the usage in different languages. 
This includes models such as 
German BERT and Italian BERT by DMBMDZ\footnote{https://github.com/dbmdz/berts} and deepset\footnote{https://www.deepset.ai/german-bert}, 
BETO Spanish BERT \cite{CaneteCFP2020},
BERTje as a Dutch BERT \cite{devries2019bertje},
Danish BERT\footnote{https://github.com/certainly.io/nordic\_bert} by Certainly\footnote{https://certainly.io/blog/danish-bert-model/},
Swedish BERT\footnote{https://huggingface.co/KB/bert-base-swedish-cased} by the National Library of Sweden / KLab,
BERT for Finnish \cite{virtanen2019multilingual},
Polbert for Polish\footnote{https://huggingface.co/dkleczek/bert-base-polish-cased-v1} by Darek Kłeczek,
BERTimbau for Portuguese \cite{souza2020bertimbau},
and Japanese BERT\footnote{https://github.com/cl-tohoku/bert-japanese/tree/v1.0} by Tohoku University. 

For each jurisdiction, we evaluated a different set of BERT variants, mainly driven by the official language(s) within the respective jurisdiction.
For instance, in US jurisdictions we tested the standard BERT base variants trained on an English corpus, whereas in non-English jurisdictions we tested
language-specific variants. 
As an exception to this we have tested on all jurisdictions - except on Japan - the multilingual version of BERT\footnote{https://huggingface.co/bert-base-multilingual-uncased}, 
which is pre-trained on 102 different languages. By doing so we aim to better catch language-specific intricacies in legal names.

Another important aspect is the "casing" of the models. Some BERT variants are trained as "cased" models, whereas others are trained as "uncased".
The difference is in the way the text is handed to the model's tokenizer. In the first case, the text is tokenized as is, including any capitalized letters.
In the latter case, the text is lower--cased before training. 
We decided to test both, the cased and the uncased versions, where available. 

The commonly available BERT models are usually trained on large corpuses of text sentences with the goal of generalizing to many domains. 
Since the majority of these sentences should not be related to any legal or financial context, we came to question if a more domain specific model might lead to better performance.
We therefore evaluated FinBERT \cite{huang2023finbert, yang2020finbert}, which has been specifically pre-trained
for the financial domain on corporate 10-K \& 10-Q ports, as well as earning call transcripts and analyst reports. 

Regardless of the specific model variation or underlying jurisdiction data, we optimized each model 
for 5 epochs with an AdamW optimizer as described in \cite{loshchilov2017decoupled}, with a learning rate $\gamma = 0.00002$ and a weight decay of $\lambda = 0.01$.

%% file: 04_experiments.tex
\section{Evaluation}

In this section, we explain our experimental setup, evaluate classifcation performance of Traditional and Transformer approaches, 
and present various findings related to entity legal form classification.

\input{041_table_comparison.tex}

\textbf{LEI dataset preparation:}
We have evaluated the presented approaches on a subset of the publicly available LEI data\footnote{Downloaded as 2022/09/14 8:00 GMT from https://www.gleif.org/en/lei-data/gleif-golden-copy/download-the-golden-copy.}, 
containing over 1.1 million legal entities from the 30 largest jurisdictions measured by their total number of entities\footnote{We did not include China because their class distribution is heavily skewed towards the ELF code \texttt{ECAK} (\begin{CJK}{UTF8}{bsmi}企业\end{CJK}) with 97.8\% of all its entities. Also, we did not include Canada due to difficulties in resolving legal forms on state and on sub-division level.}.
We prepared a separate dataset for each jurisdiction\footnote{Based on the field Entity.LegalJurisdiction}, 
including the entity legal name as well as the ELF code. 
We only selected legal entities that are active and have recently been revalidated\footnotemark[9].
This ensures that the ELF codes assigned to the entities utilize the latest version of the ELF code list.

The number of samples per jurisdiction varies from a minimum of 4,836 entities (US-New York), a median of 23,969 entities (Belgium) and a maximum of 135,079 samples (Germany). 
The number of class labels varies, ranging from a minimum of nine ELF codes (British Virgin Islands) to a median of 21 ELF codes (Austria) and a maximum of 165 ELF codes (France)\footnote{This does not necessarily include all defined ELF codes for the respective jurisdictions, but only those assigned to any entity within the data in scope.}.

Table \ref{table:comparison} shows the results as a comparison of Traditional and Transformer-based models. 
We chose to evaluate our model in terms of F1 score
to account for precision and recall simultaneously. 
Additionally, we look at Macro F1 Score (F1-M), which is useful for multiclass classification problems where classes are imbalanced. 
Each model has been cross-validated with 5 stratified, non-overlapping folds. The scores have been computed on the concatenated predictions of all folds.

\textbf{General findings:}
As a first remark, the classification performance varies a lot in between jurisdictions, which emphasizes the uniqueness of each jurisdiction in terms of the challenges it
poses to the classification task. These challenges include the varying number of samples for training, and even more the varying number of target classes and class distribution balance.
Not at last though, there are jurisdiction-specific intriciacies which have impact on the quality of the data and therefore influence the performance.    
Notably, Belgium (BE) and France (FR) exhibit significantly lower scores than other jurisdictions. In Belgium and France we generally observe that the majority of entities do not carry any legal 
form information within their legal names. Furthermore, France comes with an exceptionally high number of 165 ELF codes as target classes. 

\textbf{Traditional vs Transformer:}
Considering Traditional methods only, we found Random Forest with pre--processing to perform best in 17 jurisdictions for F1 score,
while for the Macro F1 score, a Decision Tree Classifier with pre--processing turned out to be the best in 13 jurisdictions, followed by RF + prep in 9 jurisdictions.
When comparing both methodologies, we see that Transformers outperform the Traditional pipeline in 22 out of 30 cases in the F1 score. 
On the Macro F1 score, we find that the Transformers still outperform in 9 cases. Its strongest competitor here is the Decision Tree (with and without pre--processing), 
to which it loses 15 times on the F1-M score. This can be explained by a tendency of the Transformer to being beneficial for the majority classes, while being comparably less accurate 
to classify weakly represented classes with just a few samples.

\textbf{The effect of pre--processing:}
For the Traditional models, we evaluated the impact of adding pre--processing to the general performance of the pipeline. 
We compared for each classification method the number of instances in which the pre--processing enabled pipeline outperforms its simpler counterpart. 
For the 30 jurisdictions and for both scores (in total: 60 cases), incorporating the pre--processing was superior to omitting it in 32 (CNB + prep), 45 (DT + prep), 48 (RF + prep), and 46 (SVM + prep) cases respectively.
This strongly suggests that the presented pre-processing is indeed supportive for the task within the traditional setup. 

\textbf{The effect of language:}
Looking at the Transformer models, it turned out that the multilingual BERT version performs best in 18 jurisdictions.
This is mostly due the fact that in many cases we have not found suitable language-specific models for fine-tuning. 
However, for jurisdictions like Switzerland and Luxembourg, which have multiple administrative languages, or Liechtenstein, which exhibits a large amount of multi-language names, 
using a fine-tuned multilingual BERT version performed better than a fine-tuned German BERT. 
Also, the multilingual model for example slightly outperformed BERTje (F1 0.9834, F1-M 0.7582) in the Netherlands.
For some jurisdictions though, we were able to find language-specific models that outperformed the multilingual version, including
Germany, Italy, Denmark, Finland, Austria, Poland and Japan. 
Our results indicate that language-specific models are often preferrable, but
it is fair and sometimes beneficial to use a multilingual model as a default.

In the case of Japan, the Transformer clearly outperformed the Traditional approaches. One reason being that no delimiters in the Japanese language exist to tokenize at \cite{hirabayashi2020composing}, which causes problems to the Bag-of-Words approach. 
The Japanese Bert developed by Tohoku University, though, uses a custom Japanese Tokenizer, and thus is able to achieve a high F1 score of 0.9828.

\textbf{The effect of capitalization:}
Regarding the capitalization of the models, we found that throughout all jurisdictions, the uncased variants perform better than the cased variants. 
This may be attributed to the fact that many entities are reported in a fully upper-cased fashion. The portion of upper-cased legal names varies across all jurisdictions.
A cased model, which is relying on the input text using upper and lower case characters, might be negatively affected by fully upper-cased entity names, whereas
an uncased variant might be more robust against such deficiencies in the data.

\textbf{The effect of domain specific pre-training:}
For the subset of mainly English speaking jurisdictions (GB, US-DE, US-MA, US-CA, US-NY, AU, IE, VG, KY) we evaluated FinBERT.
Even though the performance was mostly close, FinBERT, in general, was not able to outperform the standard bert-base and multilingual models.
The only exception is US-Delaware, for which it was on par with bert-base-uncased in the F1 score, and even outperformed it at the F1-M score (0.5719 vs. 0.5248).

\textbf{Learning of non-obvious relationships:}
There are cases of entity types for which there exist no explicit rules to include legal form information within a legal name. 
However, presuming sufficient domain knowledge, the legal form can sometimes be determined from non-obvious name characteristics. 
For instance, the legal entity "Langholtgaard" is registered with ELF code "Enkeltmandsvirksomhed" (\texttt{FUKI}). 
The name indicates that the entity is a farm. It turns out that in Denmark, the legal form of "Enkeltmandsvirksomhed" (english: "sole proprietor")
is predominantly used for farms. These kinds of relationships are hard to cover with rule-based approaches. 
The presented techniques are generally able to statistically model such patterns and thus acquire non-obvious domain expertise.

\textbf{Importance of sequence modelling:}
In our analysis, we found that detecting legal forms accurately sometimes requires considering the order of tokens in the entity names.
We observed several cases where the transformer model successfully identified the legal form by taking the token order into account.
However, we did not quantify this behavior in our study. 

To illustrate the point, let's consider two German entities:
Entity A has the legal name "Unsere Kinder, unsere Zukunft – Stiftung der Volksbank Odenwald eG",
while entity B has the name "Volksbank Odenwald eG". 
Both names exhibit the abbreviation "eG", which represents the legal form "eingetragene Genossenschaft" (ELF code \texttt{AZFE}). 
This would be correct for entity B, however, entity A represents a "Stiftung" (ELF code \texttt{V2YH}) (english: foundation) that belongs to entity B.

We found Bag-of-Words to not classify correctly in this scenario, as it in general cannot decide which token has more relevance - "Stiftung" or "eG". 
Also, we claim that any simplistic rule-based approach will have problems here. 
The Transformer however was able to classify this correctly.

Using the python library \textit{transformers-interpret}\footnote{https://github.com/cdpierse/transformers-interpret},
which utilizes \textit{integrated gradients} \cite{sundararajan2017axiomatic}, we are able to compute and visualize in Table \ref{table:tokensequence}
the relevance attribution of each token to the Transformer's entity legal form prediction.  
Please note that the legal name of entity B can be understood as substring of entity A's legal name.

\begin{table}[h]

\caption{Consideration of token sequence}
\begin{tabular}{lccr}
\multicolumn{1}{c}{\textbf{}} & \multicolumn{1}{c}{\textbf{Entity A}} & \multicolumn{1}{c}{\textbf{Entity B}}   \\
\toprule
Token & Attribution & Attribution\\
\midrule

[CLS] & 0.00 & 0.00 \\

unsere & 0.11 & \\
kinder & 0.12 & \\
, & 0.05 & \\
unsere & 0.10 & \\
zukunft & 0.17 & & \\
- & 0.08 & & \\
stiftung & \cellcolor[gray]{0.6} 0.82 & \\
der & \cellcolor[gray]{0.6} 0.43 & \\
volksbank & 0.01 & 0.09 \\
oden & 0.06 & 0.06\\
\#\#wald & 0.18 & 0.17\\
eg & 0.18 & \cellcolor[gray]{0.6} 0.98\\

[SEP] & 0.00 & 0.00\\
\midrule
Prediction: & \texttt{V2YH} \checkmark & \texttt{AZFE} \checkmark \\
\bottomrule
\label{table:tokensequence}
\end{tabular}
\end{table}

For entity name A, the transformer attributes the highest scores to the tokens "stiftung" and "der", 
which represent the phrase "foundation of", resulting in a correct classification of the legal form. 
For entity name B, the highest attribution is correctly associated with the token "eg". 
This demonstrates how the transformer model is able to correctly predict the legal form for both entities due to its ability to
consider the positioning and sequence of individual tokens.

\section{Expert review}

In order to validate the plausibility of the models' results, we selected ten jurisdictions with models having a high F1-score. 
In these jurisdictions, we examined entities, for which the model predicted an ELF code different from the recorded ELF code 
present in the LEI data at that time. Upon manual review, we found that many of the model's predictions were indeed correct, 
especially for instances with a high prediction probability value. 

To further evaluate our findings we utilized GLEIF's challenge facility\footnote{https://www.gleif.org/en/lei-data/gleif-data-quality-management/challenge-lei-data} 
and requested updates to the LEI data based on the legal forms suggested by our models.
The suggested legal forms were then reviewed by the respective local LEI issuing organizations, which are considered 
experts in entity data management within their accredited jurisdictions. 
A detailed view on how many of these requests have been accepted can be found 
in Table \ref{table:expertreview}.

In total, we requested updates for 7,256 entities in the ten selected jurisdictions, and
6,088 of these entities were updated based on the proposed entity legal form, resulting in an overall acceptance rate of 83.9\%.
The high acceptance rate proves that our models produce plausible legal form classifications.
Moreover, this indicates that the vast majority of LEI data accurately reflects 
the correct legal form information and therefore serves as a reliable source for training.

\begin{table}[h]

\caption{Requested reference data updates}
\begin{tabular}{lp{1.5cm}p{1.5cm}p{1.5cm}}

\toprule
Jurisdiction & Requested updates & Entities updated & Acceptance rate \\
\midrule

DE & 3,282 & 3,224 & 98.24\% \\
ES & 1,284 & 1,273 & 99.14\% \\
DK & 750 & 317 & 42.26\% \\
GB & 547 & 420 & 76.78\% \\
LU & 395 & 253 & 64.05\% \\
CH & 324 & 169 & 52.15\% \\
US-DE & 332 & 183 & 55.12\% \\
NL & 163 & 106 & 65.03\% \\
NO & 118 & 87	& 73.72\% \\
SE & 61 & 56 & 91.8\% \\
\bottomrule

\label{table:expertreview}
\end{tabular}
\end{table}

%% file: 041_table_comparison.tex
\begin{table*}
    \caption{Comparison of Traditional and Transformer models. \textmd{For Traditional models, the best performing model is shown for both F1 and Macro F1 (F1-M) score separately. 
    For Transformers, the best BERT variant is selected solely by F1 score, but both its scores are shown.}} \label{table:comparison}
\begin{tabular}{lrr|ll|ll|lcc}
\toprule
\multicolumn{3}{c}{\textbf{LEI data}} & \multicolumn{4}{c}{\textbf{Traditional}}     & \multicolumn{3}{c}{\textbf{Transformer}} \\
Jurisdiction & \#entities & \#ELF & \multicolumn{2}{c}{Best variant by F1} & \multicolumn{2}{c|}{Best variant by F1-M} & Best variant by F1 & F1 & F1-M \\
\midrule
DE & 135,079 & 31 & RF + prep & 0.9537 & DT + prep & 0.5906 & bert-base-german-uncased & \textbf{0.9616} & \textbf{0.6174} \\
IT & 104,968 & 50 & SVC + prep & 0.8990 & DT + prep & \textbf{0.3218} & bert-base-italian-uncased & \textbf{0.9010} & 0.3121 \\
NL & 89,748 & 20 & RF + prep & 0.9812 & RF + prep & 0.7529 & bert-base-multilingual-uncased & \textbf{0.9847} & \textbf{0.7676} \\
IN & 87,491 & 35 & SVC + prep & 0.8845 & DT & \textbf{0.4872} & bert-base-multilingual-uncased & \textbf{0.8862} & 0.4705 \\
ES & 84,231 & 41 & RF + prep & 0.9491 & DT + prep & \textbf{0.5219} & bert-base-multilingual-uncased & \textbf{0.9505} & 0.5191 \\
GB & 74,847 & 29 & SVC + prep & 0.9666 & DT + prep & \textbf{0.4081} & bert-base-uncased & \textbf{0.9690} & 0.4047 \\
FR & 59,973 & 165 & SVC + prep & \textbf{0.5769} & DT & \textbf{0.1890} & bert-base-multilingual-cased & 0.5710 & 0.1107 \\
DK & 56,226 & 22 & RF + prep & 0.9349 & RF + prep & 0.5870 & danish-bert-botxo & \textbf{0.9444} & \textbf{0.5941} \\
US-DE & 54,156 & 12 & SVC & 0.9871 & RF + prep & \textbf{0.6094} & finbert-pretrain & \textbf{0.9878} & 0.5719 \\
SE & 48,083 & 18 & RF + prep & 0.9789 & RF + prep & 0.5424 & bert-base-multilingual-uncased & \textbf{0.9854} & \textbf{0.5647} \\
FI & 35,587 & 52 & RF + prep & 0.9839 & DT + prep & 0.5618 & bert-base-finnish-uncased-v1 & \textbf{0.9858} & \textbf{0.5978} \\
LU & 33,683 & 28 & SVC & 0.8565 & DT + prep & \textbf{0.4279} & bert-base-multilingual-uncased & \textbf{0.8761} & 0.3817 \\
NO & 32,996 & 27 & RF + prep & 0.9888 & DT + prep & \textbf{0.6815} & bert-base-multilingual-uncased & \textbf{0.9910} & 0.5942 \\
AT & 24,433 & 21 & RF + prep & 0.9411 & DT + prep & 0.5496 & bert-base-german-uncased & \textbf{0.9635} & \textbf{0.6001} \\
BE & 23,969 & 41 & SVC + prep & 0.5089 & DT & \textbf{0.1444} & bert-base-multilingual-uncased & \textbf{0.5344} & 0.1391 \\
KY & 20,541 & 13 & RF + prep & \textbf{0.7280} & DT + prep & \textbf{0.4627} & bert-base-multilingual-uncased & 0.7108 & 0.3844 \\
PL & 20,173 & 36 & DT + prep & \textbf{0.9898} & DT + prep & \textbf{0.6252} & bert-base-polish-uncased-v1 & 0.9879 & 0.5355 \\
AU & 15,350 & 13 & SVC + prep & \textbf{0.8887} & DT + prep & \textbf{0.3301} & bert-base-multilingual-uncased & 0.8861 & 0.3198 \\
IE & 15,294 & 19 & RF & 0.9189 & DT & \textbf{0.5116} & bert-base-uncased & \textbf{0.9251} & 0.4569 \\
VG & 15,086 & 9 & SVC + prep & \textbf{0.8663} & DT & \textbf{0.2696} & bert-base-multilingual-uncased & 0.8374 & 0.1622 \\
CZ & 14,477 & 52 & RF + prep & 0.9893 & RF + prep & \textbf{0.4355} & bert-base-multilingual-uncased & \textbf{0.9908} & 0.3824 \\
EE & 13,824 & 13 & RF + prep & 0.9954 & RF + prep & 0.6291 & bert-base-multilingual-uncased & \textbf{0.9965} & \textbf{0.6329} \\
CH & 13,742 & 28 & RF + prep & 0.9211 & RF + prep & \textbf{0.4066} & bert-base-multilingual-uncased & \textbf{0.9367} & 0.3902 \\
HU & 10,041 & 33 & RF + prep & \textbf{0.9326} & DT & \textbf{0.5791} & bert-base-multilingual-uncased & 0.9265 & 0.4511 \\
JP & 9,690 & 12 & RF + prep & 0.8968 & DT + prep & 0.2598 & bert-base-japanese & \textbf{0.9828} & \textbf{0.4400} \\
LI & 9,458 & 13 & CNB + prep & 0.9522 & CNB + prep & \textbf{0.7708} & bert-base-multilingual-uncased & \textbf{0.9525} & 0.6616 \\
US-MA & 6,987 & 13 & RF & \textbf{0.9548} & DT & \textbf{0.5107} & bert-base-multilingual-uncased & 0.9501 & 0.4969 \\
PT & 6,427 & 20 & RF + prep & \textbf{0.9129} & RF + prep & \textbf{0.2950} & bert-base-multilingual-uncased & 0.9088 & 0.2566 \\
US-CA & 6,176 & 14 & RF + prep & 0.9362 & RF + prep & \textbf{0.4067} & bert-base-uncased & \textbf{0.9399} & 0.3896 \\
US-NY & 4,836 & 10 & RF + prep & 0.9520 & DT + prep & 0.4998 & bert-base-uncased & \textbf{0.9582} & \textbf{0.5250} \\
\bottomrule
\end{tabular}
\end{table*}

%% file: 05_conclusion.tex
\section{Conclusion}

Our study successfully applied Transformer-based language models for classifying entity legal forms from raw legal entity names. 
Utilizing various BERT variants, we compared their performance against traditional baselines on a substantial subset of LEI data, 
covering over 1.1 million legal entities from 30 different jurisdictions.

Our findings revealed that the presented models can effectively learn statistical relationships that are advantageous, 
even in cases where legal entity names lack explicit representation of the corresponding legal form. 
This advantage over rule-based approaches reduces development efforts and enables scalability to multiple jurisdictions.

Furthermore, we observed that language-specific pre-training has a positive impact on the models' performance, 
underscoring the importance of considering the linguistic nuances inherent in legal names. 
The ability of language models to capture the sequential nature of legal names also proved advantageous, 
as it enhances accuracy in handling edge cases, where traditional methods may struggle.

The expert review process played a critical role in validating the reliability of our models. 
We sought confirmation from local experts in ten selected jurisdictions, 
and the vast majority of the proposed legal forms were confirmed. 
This clear indication from the experts reinforces the trustworthiness and accuracy of our legal form classifications.

The LEI data and ELF codes played a crucial role in our study, providing valuable ground truth labels for legal form classification. 
We believe that broader adoption of such standards will significantly enhance transparency, while improving data integration tasks in various domains.

In this regard, we contribute our insights and techniques as part of an open source library called LENU (Legal Entity Name Understanding)\footnote{https://github.com/Sociovestix/lenu}.
The library is freely available and is distributed under Creative Commons Zero 1.0 (CC0-1.0) Universal license\footnote{https://creativecommons.org/publicdomain/zero/1.0/}.
We invite all stakeholders to use it for entity legal form classification.

%% file: lenu.bbl

\begin{thebibliography}{27}


\ifx \showCODEN    \undefined \def \showCODEN     #1{\unskip}     \fi
\ifx \showDOI      \undefined \def \showDOI       #1{#1}\fi
\ifx \showISBNx    \undefined \def \showISBNx     #1{\unskip}     \fi
\ifx \showISBNxiii \undefined \def \showISBNxiii  #1{\unskip}     \fi
\ifx \showISSN     \undefined \def \showISSN      #1{\unskip}     \fi
\ifx \showLCCN     \undefined \def \showLCCN      #1{\unskip}     \fi
\ifx \shownote     \undefined \def \shownote      #1{#1}          \fi
\ifx \showarticletitle \undefined \def \showarticletitle #1{#1}   \fi
\ifx \showURL      \undefined \def \showURL       {\relax}        \fi
\providecommand\bibfield[2]{#2}
\providecommand\bibinfo[2]{#2}
\providecommand\natexlab[1]{#1}
\providecommand\showeprint[2][]{arXiv:#2}

\bibitem[Aggarwal and Zhai(2012)]%
        {aggarwal2012survey}
\bibfield{author}{\bibinfo{person}{Charu~C Aggarwal} {and}
  \bibinfo{person}{ChengXiang Zhai}.} \bibinfo{year}{2012}\natexlab{}.
\newblock \showarticletitle{A survey of text classification algorithms}.
\newblock \bibinfo{journal}{\emph{Mining text data}} (\bibinfo{year}{2012}),
  \bibinfo{pages}{163--222}.
\newblock


\bibitem[Breiman(2001)]%
        {breiman2001random}
\bibfield{author}{\bibinfo{person}{Leo Breiman}.}
  \bibinfo{year}{2001}\natexlab{}.
\newblock \showarticletitle{Random forests}.
\newblock \bibinfo{journal}{\emph{Machine learning}}  \bibinfo{volume}{45}
  (\bibinfo{year}{2001}), \bibinfo{pages}{5--32}.
\newblock


\bibitem[Campos and Requejo(2003)]%
        {campos2003}
\bibfield{author}{\bibinfo{person}{Carrasco~R. Campos, J.} {and}
  \bibinfo{person}{A. Requejo}.} \bibinfo{year}{2003}\natexlab{}.
\newblock \showarticletitle{Legal form and risk exposure in Spanish firms.}
\newblock \bibinfo{journal}{\emph{Spanish Economic Review}}
  \bibinfo{volume}{5} (\bibinfo{year}{2003}), \bibinfo{pages}{101--121}.
\newblock


\bibitem[Cañete et~al\mbox{.}(2020)]%
        {CaneteCFP2020}
\bibfield{author}{\bibinfo{person}{José Cañete}, \bibinfo{person}{Gabriel
  Chaperon}, \bibinfo{person}{Rodrigo Fuentes}, \bibinfo{person}{Jou-Hui Ho},
  \bibinfo{person}{Hojin Kang}, {and} \bibinfo{person}{Jorge Pérez}.}
  \bibinfo{year}{2020}\natexlab{}.
\newblock \showarticletitle{Spanish Pre-Trained BERT Model and Evaluation
  Data}. In \bibinfo{booktitle}{\emph{PML4DC at ICLR 2020}}.
\newblock


\bibitem[de~Vries et~al\mbox{.}(2019)]%
        {devries2019bertje}
\bibfield{author}{\bibinfo{person}{Wietse de Vries}, \bibinfo{person}{Andreas
  van Cranenburgh}, \bibinfo{person}{Arianna Bisazza}, \bibinfo{person}{Tommaso
  Caselli}, \bibinfo{person}{Gertjan~van Noord}, {and} \bibinfo{person}{Malvina
  Nissim}.} \bibinfo{year}{2019}\natexlab{}.
\newblock \bibinfo{title}{{BERTje}: {A} {Dutch} {BERT} {Model}}.
\newblock \bibinfo{howpublished}{arXiv:1912.09582}.
\newblock
\urldef\tempurl%
\url{http://arxiv.org/abs/1912.09582}
\showURL{%
\tempurl}


\bibitem[Devlin et~al\mbox{.}(2018)]%
        {devlin2018bert}
\bibfield{author}{\bibinfo{person}{Jacob Devlin}, \bibinfo{person}{Ming-Wei
  Chang}, \bibinfo{person}{Kenton Lee}, {and} \bibinfo{person}{Kristina
  Toutanova}.} \bibinfo{year}{2018}\natexlab{}.
\newblock \showarticletitle{Bert: Pre-training of deep bidirectional
  transformers for language understanding}.
\newblock \bibinfo{journal}{\emph{arXiv preprint arXiv:1810.04805}}
  (\bibinfo{year}{2018}).
\newblock


\bibitem[Eurostat(2010)]%
        {eurostat2010}
\bibfield{author}{\bibinfo{person}{Eurostat}.} \bibinfo{year}{2010}\natexlab{}.
\newblock \showarticletitle{Business registers. Recommendations manual}.
  \bibinfo{publisher}{Publications Office of the European Union, Luxembourg}.
\newblock
\urldef\tempurl%
\url{https://ec.europa.eu/eurostat/web/products-manuals-and-guidelines/-/ks-32-10-216}
\showURL{%
\tempurl}


\bibitem[Hirabayashi et~al\mbox{.}(2020)]%
        {hirabayashi2020composing}
\bibfield{author}{\bibinfo{person}{Teruo Hirabayashi}, \bibinfo{person}{Kanako
  Komiya}, \bibinfo{person}{Masayuki Asahara}, {and} \bibinfo{person}{Hiroyuki
  Shinnou}.} \bibinfo{year}{2020}\natexlab{}.
\newblock \showarticletitle{Composing word vectors for japanese compound words
  using bilingual word embeddings}. In \bibinfo{booktitle}{\emph{Proceedings of
  the 34th Pacific Asia Conference on Language, Information and Computation}}.
  \bibinfo{pages}{404--410}.
\newblock


\bibitem[Huang et~al\mbox{.}(2023)]%
        {huang2023finbert}
\bibfield{author}{\bibinfo{person}{Allen~H Huang}, \bibinfo{person}{Hui Wang},
  {and} \bibinfo{person}{Yi Yang}.} \bibinfo{year}{2023}\natexlab{}.
\newblock \showarticletitle{FinBERT: A large language model for extracting
  information from financial text}.
\newblock \bibinfo{journal}{\emph{Contemporary Accounting Research}}
  \bibinfo{volume}{40}, \bibinfo{number}{2} (\bibinfo{year}{2023}),
  \bibinfo{pages}{806--841}.
\newblock


\bibitem[Jensen and Meckling(1976)]%
        {jensen1976}
\bibfield{author}{\bibinfo{person}{Michael Jensen} {and}
  \bibinfo{person}{William Meckling}.} \bibinfo{year}{1976}\natexlab{}.
\newblock \showarticletitle{Theory of the firm: managerial behavior, agency
  costs, and ownership structure}.
\newblock \bibinfo{journal}{\emph{Journal of Financial Economics}}
  \bibinfo{volume}{3}, \bibinfo{number}{4} (\bibinfo{year}{1976}),
  \bibinfo{pages}{305--360}.
\newblock


\bibitem[Kruse et~al\mbox{.}(2021)]%
        {kruse2021developing}
\bibfield{author}{\bibinfo{person}{Felix Kruse}, \bibinfo{person}{Jan-Philipp
  Awick}, \bibinfo{person}{Jorge~Marx G{\'o}mez}, {and} \bibinfo{person}{Peter
  Loos}.} \bibinfo{year}{2021}\natexlab{}.
\newblock \showarticletitle{Developing a Legal Form Classification and
  Extraction Approach for Company Entity Matching: Benchmark of Rule-Based and
  Machine Learning Approaches}. In \bibinfo{booktitle}{\emph{Business
  Information Systems}}. \bibinfo{pages}{13--26}.
\newblock


\bibitem[Laurent(2021)]%
        {LaurentESRB}
\bibfield{author}{\bibinfo{person}{Fran{\c c}ois Laurent}.}
  \bibinfo{year}{2021}\natexlab{}.
\newblock \showarticletitle{The benefits of the Legal Entity Identifier for
  monitoring systemic risk}.
\newblock \bibinfo{journal}{\emph{ESRB Occasional Paper Series}}
  \bibinfo{number}{18} (\bibinfo{year}{2021}).
\newblock


\bibitem[Loshchilov and Hutter(2017)]%
        {loshchilov2017decoupled}
\bibfield{author}{\bibinfo{person}{Ilya Loshchilov} {and}
  \bibinfo{person}{Frank Hutter}.} \bibinfo{year}{2017}\natexlab{}.
\newblock \showarticletitle{Decoupled weight decay regularization}.
\newblock \bibinfo{journal}{\emph{arXiv preprint arXiv:1711.05101}}
  (\bibinfo{year}{2017}).
\newblock


\bibitem[Loster et~al\mbox{.}(2018)]%
        {loster2018dissecting}
\bibfield{author}{\bibinfo{person}{Michael Loster}, \bibinfo{person}{Manuel
  Hegner}, \bibinfo{person}{Felix Naumann}, {and} \bibinfo{person}{Ulf Leser}.}
  \bibinfo{year}{2018}\natexlab{}.
\newblock \showarticletitle{Dissecting Company Names using Sequence Labeling.}.
  In \bibinfo{booktitle}{\emph{LWDA}}. \bibinfo{pages}{227--238}.
\newblock


\bibitem[Magerman et~al\mbox{.}(2006)]%
        {magerman2006data}
\bibfield{author}{\bibinfo{person}{Tom Magerman}, \bibinfo{person}{Bart
  Van~Looy}, {and} \bibinfo{person}{Xiaoyan Song}.}
  \bibinfo{year}{2006}\natexlab{}.
\newblock \showarticletitle{Data production methods for harmonized patent
  statistics: Patentee name harmonization}.
\newblock  (\bibinfo{year}{2006}).
\newblock


\bibitem[Magnani and Montesi(2007)]%
        {magnani2007study}
\bibfield{author}{\bibinfo{person}{Matteo Magnani} {and}
  \bibinfo{person}{Danilo Montesi}.} \bibinfo{year}{2007}\natexlab{}.
\newblock \showarticletitle{A study on company name matching for database
  integration}.
\newblock \bibinfo{journal}{\emph{Technical Report UBLCS-07-15, Department of
  Computer Science University of Bologna}} (\bibinfo{year}{2007}).
\newblock


\bibitem[Minaee et~al\mbox{.}(2021)]%
        {minaee2021deep}
\bibfield{author}{\bibinfo{person}{Shervin Minaee}, \bibinfo{person}{Nal
  Kalchbrenner}, \bibinfo{person}{Erik Cambria}, \bibinfo{person}{Narjes
  Nikzad}, \bibinfo{person}{Meysam Chenaghlu}, {and} \bibinfo{person}{Jianfeng
  Gao}.} \bibinfo{year}{2021}\natexlab{}.
\newblock \showarticletitle{Deep learning--based text classification: a
  comprehensive review}.
\newblock \bibinfo{journal}{\emph{ACM computing surveys (CSUR)}}
  \bibinfo{volume}{54}, \bibinfo{number}{3} (\bibinfo{year}{2021}),
  \bibinfo{pages}{1--40}.
\newblock


\bibitem[Rennie et~al\mbox{.}(2003)]%
        {rennie2003tackling}
\bibfield{author}{\bibinfo{person}{Jason~D Rennie}, \bibinfo{person}{Lawrence
  Shih}, \bibinfo{person}{Jaime Teevan}, {and} \bibinfo{person}{David~R
  Karger}.} \bibinfo{year}{2003}\natexlab{}.
\newblock \showarticletitle{Tackling the poor assumptions of naive bayes text
  classifiers}. In \bibinfo{booktitle}{\emph{Proceedings of the 20th
  international conference on machine learning (ICML-03)}}.
  \bibinfo{pages}{616--623}.
\newblock


\bibitem[Schild et~al\mbox{.}(2017)]%
        {schild10f}
\bibfield{author}{\bibinfo{person}{Christopher-Johannes Schild},
  \bibinfo{person}{Simone Schultz}, {and} \bibinfo{person}{Franco Wieser}.}
  \bibinfo{year}{2017}\natexlab{}.
\newblock \showarticletitle{Linking Deutsche Bundesbank Company Data using
  Machine-Learning-Based Classification}. Technical Report 2017-01, Deutsche
  Bundesbank Research Data and Service Centre.
\newblock


\bibitem[Sebastiani(2002)]%
        {sebastiani2002machine}
\bibfield{author}{\bibinfo{person}{Fabrizio Sebastiani}.}
  \bibinfo{year}{2002}\natexlab{}.
\newblock \showarticletitle{Machine learning in automated text categorization}.
\newblock \bibinfo{journal}{\emph{ACM computing surveys (CSUR)}}
  \bibinfo{volume}{34}, \bibinfo{number}{1} (\bibinfo{year}{2002}),
  \bibinfo{pages}{1--47}.
\newblock


\bibitem[Souza et~al\mbox{.}(2020)]%
        {souza2020bertimbau}
\bibfield{author}{\bibinfo{person}{F{\'a}bio Souza}, \bibinfo{person}{Rodrigo
  Nogueira}, {and} \bibinfo{person}{Roberto Lotufo}.}
  \bibinfo{year}{2020}\natexlab{}.
\newblock \showarticletitle{{BERT}imbau: pretrained {BERT} models for
  {B}razilian {P}ortuguese}. In \bibinfo{booktitle}{\emph{9th Brazilian
  Conference on Intelligent Systems, {BRACIS}, Rio Grande do Sul, Brazil,
  October 20-23 (to appear)}}.
\newblock


\bibitem[Sundararajan et~al\mbox{.}(2017)]%
        {sundararajan2017axiomatic}
\bibfield{author}{\bibinfo{person}{Mukund Sundararajan}, \bibinfo{person}{Ankur
  Taly}, {and} \bibinfo{person}{Qiqi Yan}.} \bibinfo{year}{2017}\natexlab{}.
\newblock \showarticletitle{Axiomatic attribution for deep networks}. In
  \bibinfo{booktitle}{\emph{International conference on machine learning}}.
  PMLR, \bibinfo{pages}{3319--3328}.
\newblock


\bibitem[Vaswani et~al\mbox{.}(2017)]%
        {vaswani2017attention}
\bibfield{author}{\bibinfo{person}{Ashish Vaswani}, \bibinfo{person}{Noam
  Shazeer}, \bibinfo{person}{Niki Parmar}, \bibinfo{person}{Jakob Uszkoreit},
  \bibinfo{person}{Llion Jones}, \bibinfo{person}{Aidan~N Gomez},
  \bibinfo{person}{{\L}ukasz Kaiser}, {and} \bibinfo{person}{Illia
  Polosukhin}.} \bibinfo{year}{2017}\natexlab{}.
\newblock \showarticletitle{Attention is all you need}.
\newblock \bibinfo{journal}{\emph{Advances in neural information processing
  systems}}  \bibinfo{volume}{30} (\bibinfo{year}{2017}).
\newblock


\bibitem[Virtanen et~al\mbox{.}(2019)]%
        {virtanen2019multilingual}
\bibfield{author}{\bibinfo{person}{Antti Virtanen}, \bibinfo{person}{Jenna
  Kanerva}, \bibinfo{person}{Rami Ilo}, \bibinfo{person}{Jouni Luoma},
  \bibinfo{person}{Juhani Luotolahti}, \bibinfo{person}{Tapio Salakoski},
  \bibinfo{person}{Filip Ginter}, {and} \bibinfo{person}{Sampo Pyysalo}.}
  \bibinfo{year}{2019}\natexlab{}.
\newblock \showarticletitle{Multilingual is not enough: BERT for Finnish}.
\newblock \bibinfo{journal}{\emph{arXiv preprint arXiv:1912.07076}}
  (\bibinfo{year}{2019}).
\newblock


\bibitem[Wolf et~al\mbox{.}(2019)]%
        {wolf2019huggingface}
\bibfield{author}{\bibinfo{person}{Thomas Wolf}, \bibinfo{person}{Lysandre
  Debut}, \bibinfo{person}{Victor Sanh}, \bibinfo{person}{Julien Chaumond},
  \bibinfo{person}{Clement Delangue}, \bibinfo{person}{Anthony Moi},
  \bibinfo{person}{Pierric Cistac}, \bibinfo{person}{Tim Rault},
  \bibinfo{person}{R{\'e}mi Louf}, \bibinfo{person}{Morgan Funtowicz},
  {et~al\mbox{.}}} \bibinfo{year}{2019}\natexlab{}.
\newblock \showarticletitle{Huggingface's transformers: State-of-the-art
  natural language processing}.
\newblock \bibinfo{journal}{\emph{arXiv preprint arXiv:1910.03771}}
  (\bibinfo{year}{2019}).
\newblock


\bibitem[Wu et~al\mbox{.}(2016)]%
        {wu2016google}
\bibfield{author}{\bibinfo{person}{Yonghui Wu}, \bibinfo{person}{Mike
  Schuster}, \bibinfo{person}{Zhifeng Chen}, \bibinfo{person}{Quoc~V Le},
  \bibinfo{person}{Mohammad Norouzi}, \bibinfo{person}{Wolfgang Macherey},
  \bibinfo{person}{Maxim Krikun}, \bibinfo{person}{Yuan Cao},
  \bibinfo{person}{Qin Gao}, \bibinfo{person}{Klaus Macherey}, {et~al\mbox{.}}}
  \bibinfo{year}{2016}\natexlab{}.
\newblock \showarticletitle{Google's neural machine translation system:
  Bridging the gap between human and machine translation}.
\newblock \bibinfo{journal}{\emph{arXiv preprint arXiv:1609.08144}}
  (\bibinfo{year}{2016}).
\newblock


\bibitem[Yang et~al\mbox{.}(2020)]%
        {yang2020finbert}
\bibfield{author}{\bibinfo{person}{Yi Yang}, \bibinfo{person}{Mark
  Christopher~Siy Uy}, {and} \bibinfo{person}{Allen Huang}.}
  \bibinfo{year}{2020}\natexlab{}.
\newblock \showarticletitle{Finbert: A pretrained language model for financial
  communications}.
\newblock \bibinfo{journal}{\emph{arXiv preprint arXiv:2006.08097}}
  (\bibinfo{year}{2020}).
\newblock


\end{thebibliography}
